\documentclass{article}

    \PassOptionsToPackage{numbers, compress}{natbib}

    \usepackage[final]{neurips_data_2021}

\usepackage[utf8]{inputenc} 
\usepackage[T1]{fontenc}    
\usepackage{hyperref}       
\usepackage{url}            
\usepackage{booktabs}       
\usepackage{amsfonts}       
\usepackage{nicefrac}       
\usepackage{microtype}      
\usepackage{xcolor}         

\usepackage{multirow}
\usepackage{CJKutf8}
\usepackage{microtype}
\usepackage{makecell}
\usepackage{graphicx}

\newcommand{\tabincell}[2]{\begin{tabular}{@{}#1@{}}#2\end{tabular}}  

\newcommand{\name}{{\textrm{NCR}}}
\newcommand{\Cs}{{\textrm{$\mathrm{C^3}$}}}
\newcommand{\CsD}{{\textrm{$\mathrm{C^3_D}$}}}
\newcommand{\CsM}{{\textrm{$\mathrm{C^3_M}$}}}

\newcommand{\xss}[1]{{#1}} 

\title{\emph{Native Chinese Reader}: A Dataset Towards \\Native-Level Chinese Machine Reading Comprehension}

\author{%
Shusheng Xu$^{1}$\thanks{Equal contribution.}\,,\, Yichen Liu$^{2,5}{}^*$, Xiaoyu Yi$^{3,5}$, Siyuan Zhou$^{4,5}$, Huizi Li$^{5}$, Yi Wu$^{1, 6}$\\[1ex]
	$^1$ IIIS, Tsinghua University, $^2$ New York University
	$^3$ Shenzhen University,  \\
	$^4$ Peking University, $^5$ Haihua Institute for Frontier Information Technology,\\ $^6$ Shanghai Qi Zhi Institute \\[1ex]
	{\tt xuss20@mails.tsinghua.edu.cn}, {\tt yl7043@nyu.edu}, {\tt yixiaoyuszu@outlook.com} \\
	{ \tt elegyhunter@gmail.com}, {\tt lhz@haihua.org.cn},  {\tt jxwuyi@gmail.com}\\ 
}

\begin{document}

\maketitle
\begin{abstract}
We present \emph{Native Chinese Reader (\name)}, a new machine reading comprehension (MRC) dataset with particularly long articles in both \emph{modern} and \emph{classical} Chinese. {\name} is collected from the exam questions for the Chinese course in China's high schools, which are designed to evaluate the language proficiency of native Chinese youth. 
Existing Chinese MRC datasets are either domain-specific or focusing on short contexts of a few hundreds of characters  in modern Chinese only.
By contrast, {\name} contains 8390 documents with an average length of 1024 characters covering a wide range of Chinese writing styles, including modern articles, classical literature and classical poetry.
A total of 20477 questions on these documents also require strong reasoning abilities and common sense  to figure out the correct answers. 
We implemented multiple baseline models using popular Chinese pretrained models and additionally launched an online competition using our dataset to examine the limit of current methods. The best model achieves 59\% test accuracy while human evaluation shows an average accuracy of 79\%, which indicates a significant performance gap between current MRC models and native Chinese speakers. We release the dataset at \url{https://sites.google.com/view/native-chinese-reader/}.

\end{abstract}

\section{Introduction}


Machine reading comprehension (MRC) is one of the fundamental tasks in natural language understanding, which requires a machine to read a document to correctly answer questions based on the context. MRC  has attracted significant efforts from both academia and industry with continuous development of MRC datasets~\cite{CNNDM,MSMARCO,RACE,SQuAD2,CoQA,DREAM}, which also keeps pushing the frontier of MRC models and learning algorithms~\cite{BERT,Roberta,Albert,GPT3} to eventually bridge the gap between AI systems and human readers. 

In addition to the advances in English MRC, researchers have also made substantial progresses in Chinese MRC challenges with many high-quality Chinese MRC datasets released. These MRC datasets focus on a variety of domains of Chinese understanding, such as fact extraction~\cite{CMRC2017,CMRC2018}, dialogue understanding~\cite{C3}, common sense~\cite{DuReader}, idiom selection~\cite{ChID} and exam questions used in language proficiency tests~\cite{C3}.

However, all these datasets provide particularly limited challenges for the purpose of building MRC models with the same language proficiency as \emph{native} Chinese speakers. There are 3 major dataset limitations. First, the length of reading materials are \textbf{\emph{short}}. For example, {\CsM}~\cite{C3}, a multiple-choice MRC dataset with the longest documents, has merely 180 characters per document on average. Even in the cloze-based datasets, the longest average document length is just around 500 characters. Second, the questions are \textbf{\emph{not sufficiently difficult}}. Most existing datasets are either extractive or domain-specific (e.g., focusing on idiom or simple facts). Although {\Cs}~\cite{C3} provides exam-based free-form multiple-choice questions, they are designed for \emph{non-native} speakers and therefore do not require native-level reasoning capabilities and common sense knowledge to answer the questions. More importantly, none of existing datasets consider reading comprehension on \textbf{\emph{classical Chinese}} documents, such as classical literature and poetry. Classical Chinese, as a writing style used in almost all formal writing until early 20th century~\cite{wiki}, plays a critical role in Chinese culture and has led to numerous idioms and proverbs. Even today, classical literature and poetry are still widely taught and examined in China's education system.

We developed a new general-form multiple-choice Chinese MRC dataset, \emph{Native Chinese Reader (\name)}, towards building a \emph{native-level} Chinese comprehension system. 
The {\name} dataset contains 8390 documents with over 20K questions collected using the exam questions for the Chinese course in China's high schools, which are designed to evaluate the language proficiency of \emph{native} Chinese youth. Therefore, {\name} naturally overcomes the limitations of existing datasets with sufficiently challenging questions and long documents in an average length of 1024 characters over both modern and classical Chinese writing styles (see Table~\ref{tab:comparison}).

We provided in-depth analysis of {\name} and implemented baselines using popular pretrained models. To further examine the limit of current MRC methods, we additionally launched a  online competition using {\name}. 
The best model we obtained achieves an average test accuracy of 59\%, which is far below human evaluation result of 79\% accuracy. This suggests a significant gap between current MRC model capabilities and the native-level Chinese language proficiency. We hope that {\name} could serve as a milestone for the community to benefit future breakthroughs in Chinese natural language understanding.

\section{Related Work}


\paragraph{English Datasets:}
Machine Reading Comprehension tasks require a machine to answer a question based on the content in the given document. Early MRC datasets are primarily \emph{cloze/span-based}, where the answer is simply a span in the document or a few words to be filled in the blank, including CNN/Daily Mail~\cite{CNNDM}, LAMBADA~\cite{LAMBADA}, CBT\cite{CBT}, BookTest~\cite{BookTest}, Who-did-What~\cite{Who-did-what} and CLOTH~\cite{CLOTH}.
The famous SQuAD dataset~\cite{SQuAD1,SQuAD2} for the first time introduces human-generated \emph{free-form questions}, which requires the machine to understand natural language to select the correct span in Wikipedia pages. Similar datasets follow this trend of using free-form questions and adopt reading documents from a variety of sources, such as news articles~\cite{NewsQA,TriviaQA} and dialogues~\cite{NarrativeQA,CoQA,QuAC}.
In addition to these datasets where the answers can be directly \emph{extracted} from the document, another popular type of datasets, i.e., \emph{abstractive} datasets, ask the reader to generate an answer that may not be found in the given context~\cite{MSMARCO,DROP}. Abstractive datasets further require the reader to perform non-trivial reasoning over the facts in the document as well as common sense knowledge to produce answers. However, since the answer itself is a natural language, evaluation for abstractive datasets can be tricky.
\emph{Multiple-choice} datasets overcome the evaluation difficulty in abstractive datasets by simply asking the reader to select the correct answer from the candidate options. Representative datasets, such as RACE~\cite{RACE} and DREAM~\cite{DREAM}, utilize exam questions collected from standard English proficiency tests, which are generated by language teachers to evaluate a variety of language capabilities of non-native English speakers.

\paragraph{Chinese Datasets:}
The development of Chinese MRC datasets follow a similar trend of English ones. Early cloze-based datasets, such as People Daily news (PD) dataset and Children’s Fairy Tale (CFT) dataset~\cite{PDCFT},  utilize a sentence with a repeated noun removed as the question and ask the reader to predict the removed noun. As Chinese counterparts to the SQuAD dataset, DRCD~\cite{DRCD}, CMRC2017~\cite{CMRC2017} and CMRC2018~\cite{CMRC2018} datasets adopt human-generated questions and ask the reader to extract spans in the given documents as answers. DuReader~\cite{DuReader}, a representative abstractive dataset collects natural questions and answers from Baidu search data, which are in the same style as the English MS-MARCO dataset~\cite{MSMARCO}. 
CMRC2019 dataset~\cite{CMRC2019} and ChID dataset~\cite{ChID} combine cloze-based questions and multiple-choice answer options. In CMRC2019, a few sentences are masked in each document and the reader is asked to match each option sentence to the corresponding blank in the document. ChID focuses on traditional Chinese idioms by asking the reader to select the correct idiom based on the given story context. 
Recently, the {\Cs} dataset~\cite{C3} was released, which contains both free-form questions and multiple-choice answer options. {\Cs} is collected using the exam questions for Chinese-as-a-second-language tests and consists of two sub-datasets, {\CsD} focusing on normal documents and {\CsM} on dialogues, which can be viewed as the Chinese counterparts of RACE~\cite{RACE} and DREAM~\cite{DREAM} respectively.
\begin{table}[bt]
    \caption{Comparison between {\name} and related Chinese MRC datasets. NQ is short for free-form \emph{natural question}.}
    \label{tab:comparison}
    \centering
    \small
    \begin{tabular}{c|cccccc}
    \hline
Dataset & \#Que. & Source of Doc. & Que. type & Ans. Type & \tabincell{c}{Doc. \# Token \\ Avg.} & \tabincell{c}{Classical \\Chinese}\\
    \hline
        PD & 877K & News & cloze & extractive & 379 & No\\
        CFT & 3.5K & Stories & cloze & extractive & 139 & No\\
        DRCD & 34K & Wiki & NQ & extractive & 437 & No\\
        CMRC 2017 & 364K & Wiki & NQ & extractive & 486 & No\\
        CMRC 2018 & 18K & Wiki & NQ & extractive & 508 & No\\
        DuReader & 200K  & Baidu & NQ & abstractive & 82 & No\\
        CMRC 2019 & 100K & Story & cloze & multiple choice & 557 & No\\
        ChID & 729K & News\&Stories & cloze & multiple choice & 159 & No\\
        C3-D & 9.6K & Exam \scriptsize(Non-Native)& NQ & multiple choice & 76 & No\\
        C3-M & 10K & Exam \scriptsize(Non-Native)& NQ & multiple choice & 180 & No\\
    \hline
        NCR & 20.4K & Exam \scriptsize(Native) & NQ & multiple choice & 1024 & Included\\
    \hline
    \end{tabular}
\end{table}
\paragraph{Position of {\name}:}
We developed Native Chinese Reader (\name), a exam-question-based MRC dataset with free-from questions and multiple-choice answer options, which aims to push the frontier of building \emph{native-level} Chinese MRC models. 
The high-level statistics of {\name} and all the aforementioned datasets are summarized in Table.~\ref{tab:comparison}.
{\Cs} is perhaps the most related work to ours. However, {\Cs} are collected from \emph{Chinese-as-a-second-language} tests, so its questions are much easier than {\name} for three reasons. First, documents, questions and answers in {\name} are substantially \textbf{longer} than {\Cs}. Second, a quarter of the documents in {\name} are written in \textbf{classical Chinese}, which is a critical component of Chinese language but largely ignored by existing works. We remark that although the answers in ChID dataset~\cite{ChID} are idioms, which is a restricted form of classical Chinese, the documents in ChID remain in modern Chinese. Lastly, the questions in {\name} are collected from the exams for China's high-school students and require \textbf{native-level reasoning capabilities} using the background knowledge of Chinese history and culture. In-depth comparisons on the question types between {\Cs} and {\name} can be found in Sec.~\ref{sec:question_type} with example questions shown in Table.~\ref{tab:reasoning_questions}.
\begin{CJK}{UTF8}{gkai}
\xss{In addition, we highlight that a lot of questions in \name~require choosing one \emph{incorrect} option out of 4 options (i.e., three other are \emph{correct}; see Table~\ref{tab:short_examples} and \ref{tab:reasoning_questions} for examples). We count the questions containing ``不正确'' (``incorrect''), ``不符合'' (``incompatible'') or ``不恰当'' (``inappropriate''), 56.49\%, 57.63\%, and 56.14\% of questions fall into this category in training/validation/test sets respectively. This requires the capability of understanding and reasoning with negations.
}
\end{CJK}

Finally, we remark that, in addition to Chinese and English, there are also other datasets developed in other languages like Japanese~\cite{shibuki2014overview, takahashi-etal-2019-machine}, Russian~\cite{efimov2020sberquad} and cross-lingual scenarios~\cite{artetxe2019cross, liu2019xqa}, which are of parallel interest to our project.

\section{Native Chinese Reader (\name) Dataset}
In this section, we provide detailed analysis of the documents and questions in our {\name} dataset, including overall statistics, document styles, major challenges as well as studies on question types. 


\subsection{Task Format and Collection Methodology}
\label{collection}
In {\name}, each document is associated with a series of multiple-choice questions. Each question has 2 to 4 options, of which exactly one is correct. The task is to select the correct option based on the document and the question.
Both questions and options are expressed in natural language covering a wide range of question types (more details discussed below).

All the questions and documents are collected from online open-access high-school education materials. 
After data cleaning, 8315 documents followed by \xss{20284} questions are obtained. We randomly split the dataset at the document level, with 6315 for training, 1000 for validation and 1000 for testing. 
Furthermore, to make sure our test set has sufficient novel questions that never appear online, we also invited a few high-school Chinese teachers to manually generate 193 questions for a total of 73 additional documents to augment the test set.
Finally, {\name} consists of 6315 documents with 15419 questions for training, 1000 documents with 2443 questions for validation and 1073 documents with \xss{2615} questions for testing.
\begin{table}[ht]
    \caption{The overall statistics of different Chinese multi-choice MRC datasets. ChID and CMRC2019 are cloze-based without questions. \textbf{*} means statistics are collected over validation and test set only.}
    \label{tab:length_count}
    \small
    \centering
    \begin{tabular}{l|c c c c c}
    \hline
        \textbf{Datasets} & \tabincell{c}{Len. of Doc.\\Avg./Max.} & \tabincell{c}{Len. of Que.\\Avg./Max.} & \tabincell{c}{Len. of Opt.\\Avg./Max.} &
        \tabincell{c}{\#Opt. per Que.\\Min./Avg./Max.} &
        \tabincell{c}{\#Que. per Doc.\\Min./Avg./Max.}
        \\
    \hline
        \textbf{ChID}  & 159.1 / 581 & N/A & 4 / 4 & 7 / 7.0 / 7& 1 / 1.2 / 12 \\
        \textbf{CMRC 2019} & 557.3 / 688 & N/A & 13.8 / 29 & 5 / \textbf{10.6} / 15 & 0 / \textbf{9.9} / 15\\
        \textbf{C3-M} & 180.2 / 1,274 & 13.5 / 57 & 6.5 / 45 & 2 / 3.7 / 4 & 1 / 1.9 / 6 \\
        \textbf{C3-D} & 76.3 / 1,540 & 10.9 / 34 & 4.4 / 31 & 3 / 3.8 / 4 & 1 / 1.2 / 6\\
        \textbf{C3} & 116.9 / 1,540 & 12.2 / 57 & 5.5 / 45 & 2 / 3.8 / 4 & 1 / 1.5 / 6 \\
    \hline
        \textbf{NCR} \scriptsize{Classical only}* & 521.5 / 1,258 & 25.7 / 178 & 36.8 / 130 & 2 / 4.0 / 4 & 1 / 2.2 / 5  \\
        \textbf{NCR} \scriptsize{Modern only}* & 1207.8 / 4,640 & 24.4 / 276 & 44.1 / 152 & 2 / 4.0 / 4 &  1 / 2.5 / 5 \\
        \textbf{NCR} \scriptsize{All} & \textbf{1023.7 / 4,640} & \textbf{24.5 / 352} & \textbf{43.0 / 256} & 2 / 4.0 / 4 & 1 / 2.4 / 5 \\
    \hline
    \end{tabular}
\end{table}
\subsection{Dataset Statistics}
\label{statistics}
We summarize the \xss{high-level} statistics of our {\name} dataset and other related multi-choice Chinese MCR datasets in Table~\ref{tab:length_count}. 
In addition, we also measure the statistics of classical and modern documents from the validation and test set, where we can observe that modern Chinese articles are more than twice longer than classical Chinese literature.
Comparing with other Chinese MRC datasets, {\name} is an order of magnitude longer, even including those very concise classical Chinese documents.
Besides documents, {\name} also contains much longer questions and answer options. Particularly for the option length, {\name} is almost an order of magnitude longer except the CMRC2019 dataset. We remark that CMRC2019 is a cloze-style dataset with a completely different question style from {\name}: CMRC2019 options are original document texts while the reader only needs to match the options to the corresponding blank in the document. Overall, {\name} has substantially longer articles, questions and options 
with diverse document styles, which suggests a much higher comprehension difficulty than existing datasets.



\begin{table}[ht]
    \caption{Statistics of document length over {\name} validation set and test set. Classical Chinese articles (including poetry) are much shorter than modern Chinese articles. 
    }
    \label{tab:num_clsssical}
    \centering
    \begin{tabular}{c|c|ccc}
    \hline
Style & count & min & avg & max\\
    \hline
    Modern & 1493 &  47 & 1208 & 4640 \\
    Classical & 580 & 24 & 522 & 1258 \\
    Poetry & 63 & 24 & 156 & 668 \\
    \hline
    \end{tabular}
\end{table}
\begin{table}[ht]
    \scriptsize
    \caption{Example documents and questions (left) with English translation (right). Top (\textbf{D1}): a classical Chinese poem; Bottom (\textbf{D2}): an except of a modern Chinese article. $\star$ denotes the correct option for each question (\textbf{Q}). \label{tab:short_examples}}
    \label{tab:examples}
    \begin{minipage}{1\linewidth}
    \centering
        \begin{tabular}{m{5.2cm}  m{7.9cm}}
        \toprule
        \textbf{D1} \begin{CJK}{UTF8}{gkai}
         \textbf{相见欢} \quad 李煜
        \end{CJK} & \textbf{D1}  \emph{Form of Xiang-Jian-Huan} \quad \textbf{Li Yu}\\
        \begin{CJK}{UTF8}{gkai}
        无言独上西楼，月如钩。 寂寞梧桐深院锁清秋。 剪不断，理还乱，是离愁。 别是一般滋味在心头。
        \end{CJK} &
 Silent, solitary,
I step up the western tower.
The moon appears like a hook.
The lone parasol tree locks the clear autumn in the deep courtyard.
What cannot be cut
nor raveled,
is the sorrow of separation:
Nothing tastes like that to the heart.
        \\
         \midrule
       \textbf{Q1} \begin{CJK}{UTF8}{gkai}
        “ 寂寞梧桐深院锁清秋”中“锁“的意思是
        \end{CJK} & \textbf{Q1} In "The lone parasol tree locks the clear autumn in the deep courtyard.", "Locks” means\\
        \begin{CJK}{UTF8}{gkai} \quad 
        A.锁头\quad B.金锁\end{CJK} \quad \begin{CJK}{UTF8}{gkai} \textbf{C.锁住} \end{CJK} $\star$ \quad\begin{CJK}{UTF8}{gkai} D.开锁 
        \end{CJK} & A.The lock \quad B.Gold lock \quad \textbf{C. Lock up} $\star$ \quad D.Unlock\\
        \textbf{Q2} \begin{CJK}{UTF8}{gkai}
        下面这首词的赏析不正确的一项是
        \end{CJK}
         & \textbf{Q2} The incorrect option for the appreciation and analysis of this poem is\\
        A. \begin{CJK}{UTF8}{gkai}
        上阕定景，西楼、月色、梧桐、深院、清秋，画面冷寂。
        \end{CJK} & A.The scenery is fixed in the first half, including the west tower, moonlight, parasol tree, deep courtyard, and clear autumn, the picture of which is cold and quiet.\\
        B. \begin{CJK}{UTF8}{gkai}
      “寂寞梧桐深院锁清秋”一句，写栽着梧桐树的院落很寂静，渲染了清秋气氛。
        \end{CJK} & B. The sentence ”The lone parasol tree locks the clear autumn in the deep courtyard" says that the courtyard with the parasol tree is very quiet, rendering the atmosphere of autumn.\\
        \begin{CJK}{UTF8}{gkai}
        C. 下阕转入抒怀，写出了作者隐忧生活中难以排遣的感情$\star$
        \end{CJK} & \textbf{C. The second half turns to express feelings and writes about author’s unrelievable feeling when he secretly worry about life.}$\star$\\
        D. \begin{CJK}{UTF8}{gkai}
        全词将抽象的情感加以形象化，抒发了作者离乡去国之苦。
        \end{CJK} & D. The whole poem visualizes abstract emotions and expresses the author's suffering in leaving his hometown and the capital.\\
        \bottomrule
    \end{tabular}
    \end{minipage}

    \begin{minipage}{1\linewidth}
        \centering
        \begin{tabular}{m{5.2cm} m{7.9cm}}
        \toprule
        \textbf{D2} \begin{CJK}{UTF8}{gkai}
        在酒楼上（节选）鲁迅
        ...(2) 我竟不料在这里意外的遇见朋友了，——假如他现在还许我称他为朋友。那上来的分明是我的旧同窗，也是做教员时代的旧同事，面貌虽然颇有些改变，但一见也就认识，独有行动却变得格外迂缓，很不像当年敏捷精悍的吕纬甫了。
        (11) “我一回来，就想到我可笑。”他一手擎着烟卷，一只手扶着酒杯，似笑非笑的向我说。“我在少年时，看见蜂子或蝇子停在一个地方，给什么来一吓，即刻飞去了，但是飞了一个小圈子，便又回来停在原地点，便以为这实在很可笑，也可怜。可不料现在我自己也飞回来了，不过绕了点小圈子。又不料你也回来了。你不能飞得更远些么？”
        (20) “你教的是‘子曰诗云’”么？我觉得奇异，便问。(21) “自然。你还以为教的是ABCD么？我先是两个学生，一个读《诗经》，一个读《孟子》。新近又添了一个，女的，读《女儿经》。连算学也不教，不是我不教，他们不要教。”(22) “我实在料不到你倒去教这类书,...” (23) “他们的老子要他们读这些；我是别人，无乎不可的。这些无聊的事算什么？只要随随便便,...”(24) “那么，你以后豫备怎么办呢？” (25) “以后？——我不知道。你看我们那时豫想的事可有一件如意？我现在什么也不知道，连明天怎样也不知道，连后一分...” 
        \end{CJK} &
\textbf{D2}. \emph{In the Restaurant} (Excerpt) \textbf{Lu Xun} 
...(2) I never guessed that here of all places I should
expectedly meet a friend -- if such he would still let me call him. The newcomer was an old class mate who had been my colleague when I was a teacher, and although he had changed a great deal I knew him as soon as saw him. Only he had become much slower in his movements, very unlike the nimble and active Lu Wei-fu of the old days.
(11)``As soon as I came back I knew I was a fool''. Holding
his cigarette in one hand and the wine cup in the other, he
spoke with a bitter smile. `` When I was young, I saw the way
bees or flies stopped in one place. If they were frightened
they would fly but after flying in a small circle they would
come back again to stop in the same place; and I thought this really very foolish, as well as pathetic. But I didn't think that I would fly back myself, after only flying in a small circle.
And I didn't think you would come back either. Couldn't you
have flown a little further?''     
(20)``Are you teaching that?'' I asked in astonishment
(21)``Of course. Did you think I was teaching English? First
I had two pupils, one studying the Book of Songs, the other
Mencius. Recently I have got another, a girl, who is studying
the Canon for Girls. I don't even teach mathematics; not that
I wouldn't teach it, but they don't want it taught.''
(22)``I could really never have guessed that you would be teaching such books''
(23) ``Their father wants them to study these. I'm an outsider
so it's all the same to me. Who cares about such futile affairs
anyway There's no need to take them seriously ...''
(24) ``Then what do you mean to do in future?''
(25) ``In future? I don' t know. Just think: Has any single thing
turned out as we hoped of all we planned in the past? I'm not sure of anything now, not even of what I will do tomorrow, or even of the next minute ...''
        \\
         \midrule
        \textbf{Q3} \begin{CJK}{UTF8}{gkai}
        下列对文章思想内容的理解与分析，不正确的一项是
        \end{CJK}
         &
        \textbf{Q3} The incorrect one from the following understanding and analysis of the thought content of the article is:\\
        A. \begin{CJK}{UTF8}{gkai}
        “行动却变得格外迂缓，很不像当年敏捷精悍的吕纬甫了”高度概括了眼前吕纬甫的精神状态，突出他的迂缓颓废。
        \end{CJK} &
        A. `` Only he had becomemuch slower in his movements, very unlike the nimble and active LuWei-fu of the old days. '' gives a high-level overview of Lu Weifu's mental state, highlighting his sluggish decadence.\\
        B. \begin{CJK}{UTF8}{gkai}
        \emph{“蝇子飞了一个小圈子，便又回来停在原地点”，吕纬甫的这番自述自嘲中对自身缺乏清醒的认识，浑噩度日，揭示了残酷的现实生活将人的灵魂挤扁，人们只能在颓唐消沉中磨蚀生命的主题。}$\star$ 
        \end{CJK} & 
        \textbf{B. ``If they were frightened they would fly but after flying in a small circle they would come back again to stop in the same place.'' indicates that Lu Wei-fu lacks clear understanding of himself and lives in a muddle, which reveals that the cruel reality of life squeezes the soul, human can only wear out their lives in depression.} $\star$ \\
        C. \begin{CJK}{UTF8}{gkai}
        从吕纬甫叙述现在教书生涯的内容和原因的话语中，可见他已经违背了当初的理想，变得苟且偷安，屈从于当前顽固封建势力。
        \end{CJK} & 
        C. From Lu Wei-fu's narration of the content and reasons of his current teaching career, it can be seen that he has violated his original ideals, has become stubborn, and yielded to the current stubborn feudal forces.\\
        D. \begin{CJK}{UTF8}{gkai}
        文章通过吕纬甫的人生经历来告诉读者，吕纬甫的人生悲剧正是那个时代无数知识分子悲剧命运的代表，而个人的悲剧背后则是整个时代的悲哀。 
        \end{CJK} & 
        D. The article tells readers through Lu Wei-fu's life experience that Lu Weifu's life tragedy is the representative of the tragedy of countless intellectuals in that era, and behind the personal tragedy is the tragedy of the entire era.\\
    \bottomrule
    \end{tabular}
    \end{minipage}
\end{table}
\subsection{Document Style and Challenges}
\label{article_style}

We manually annotated the writing styles of the documents in validation set and test set with summarized statistics in Table~\ref{tab:num_clsssical}. Almost a quarter of the documents are in classical Chinese. We remark that most documents are collected from online open-access resources. This indicates that classical Chinese indeed plays a critical role in China's Chinese class, which, however, is often ignored in previous Chinese MRC studies. Table~\ref{tab:short_examples} presents two example documents, one in classical Chinese (\textbf{D1}) and one in modern Chinese (\textbf{D2}), with associated questions. In the following content, we will discuss the major challenges in {\name} with respect to different document writing styles.




\paragraph{Classical Chinese: }




Classical Chinese literature is substantially more difficult than modern Chinese documents due to its conciseness and flexible grammar. 
Most classical Chinese words are expressed in a single character and therefore are not restrictively categorized into parts of speech: nouns can be used as verbs, adjectives can be used as nouns, and so on. For example, the character \begin{CJK}{UTF8}{gbsn}``东''\end{CJK} only means ``east'' in modern Chinese. However, in the classical Chinese sentence, \begin{CJK}{UTF8}{gbsn}``顺流而东也''\end{CJK}(advance eastward along the river), it actually means ``advance eastward''. 
Classical Chinese also has distinguishing sentence patterns from nowadays, such as changing the order of characters and often dropping subjects and objects when a reference to them is understood. 
Furthermore, an important sub-category in classical Chinese is \emph{poetry}, which is typified by certain traditional poetic forms and rhythms. About 10\% of the classical documents in {\name} are poetry. Table~\ref{tab:short_examples}, \textbf{D1} shows a famous classical Chinese poem from Song dynasty, which is particularly short and abstract in order to satisfy the poetic form of \begin{CJK}{UTF8}{gbsn}``相见欢''\end{CJK} (Xiang-Jian-Huan). This poem describes a scene where the poet stands on a tower staring at the moon. However, in order to correctly understand the sentiment and meaning of the poem, the reader needs to leverage imagery and symbolism in classical Chinese culture (e.g., moon and autumn mean sadness) as well as the personal background of the poet (e.g., Li Yu was a captured emperor).





\paragraph{Modern Chinese: }

For the modern Chinese documents in {\name}, in addition to the challenge due to longer average length, the associated questions also focus more on the high-level metaphors and the underlying thoughts, which often require non-trivial reasoning with historical and cultural knowledge. Table~\ref{tab:short_examples}~\textbf{D2} shows an excerpt from a long article (the full document in {\name} has about 2000 characters) written by a famous Chinese author, \begin{CJK}{UTF8}{gbsn}鲁迅\end{CJK} (Lu Xun). The article describes a scene where the author unexpectedly met one of his old friends not seen for a long time and had a meal together. The associated question (Q3) primarily asks about the high-level thoughts expressed by the author, which has to be inferred from the entire article and requires the readers to have strong knowledge of the author's personal experiences and the background of the era.

\begin{figure}[t]
    \centering
    \begin{minipage}[t]{0.45\linewidth}
    \centering
    \includegraphics[width=1\linewidth]{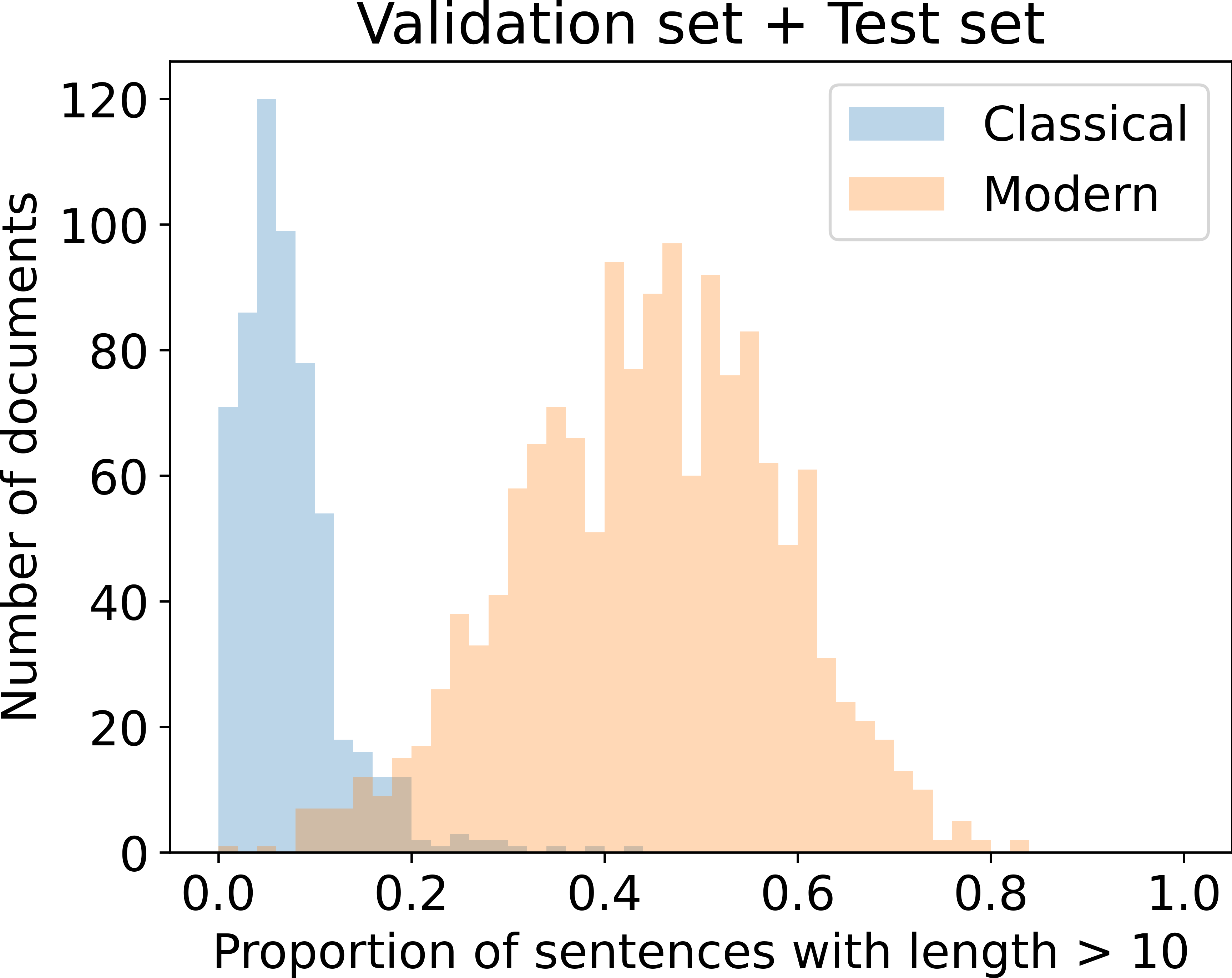}
  \caption{\xss{The histogram of $p_{(s>10)}$ over the validation and test sets. The blue bars correspond to classical documents, while the orange bars represent modern documents. We can observe that 0.2 is an effective classification boundary to distinguish classical and modern documents.}}
  \label{fig.dev+test}
    \end{minipage}
    \hspace{2mm}
    \begin{minipage}[t]{0.45\linewidth}
    \centering
    \includegraphics[width=1\linewidth]{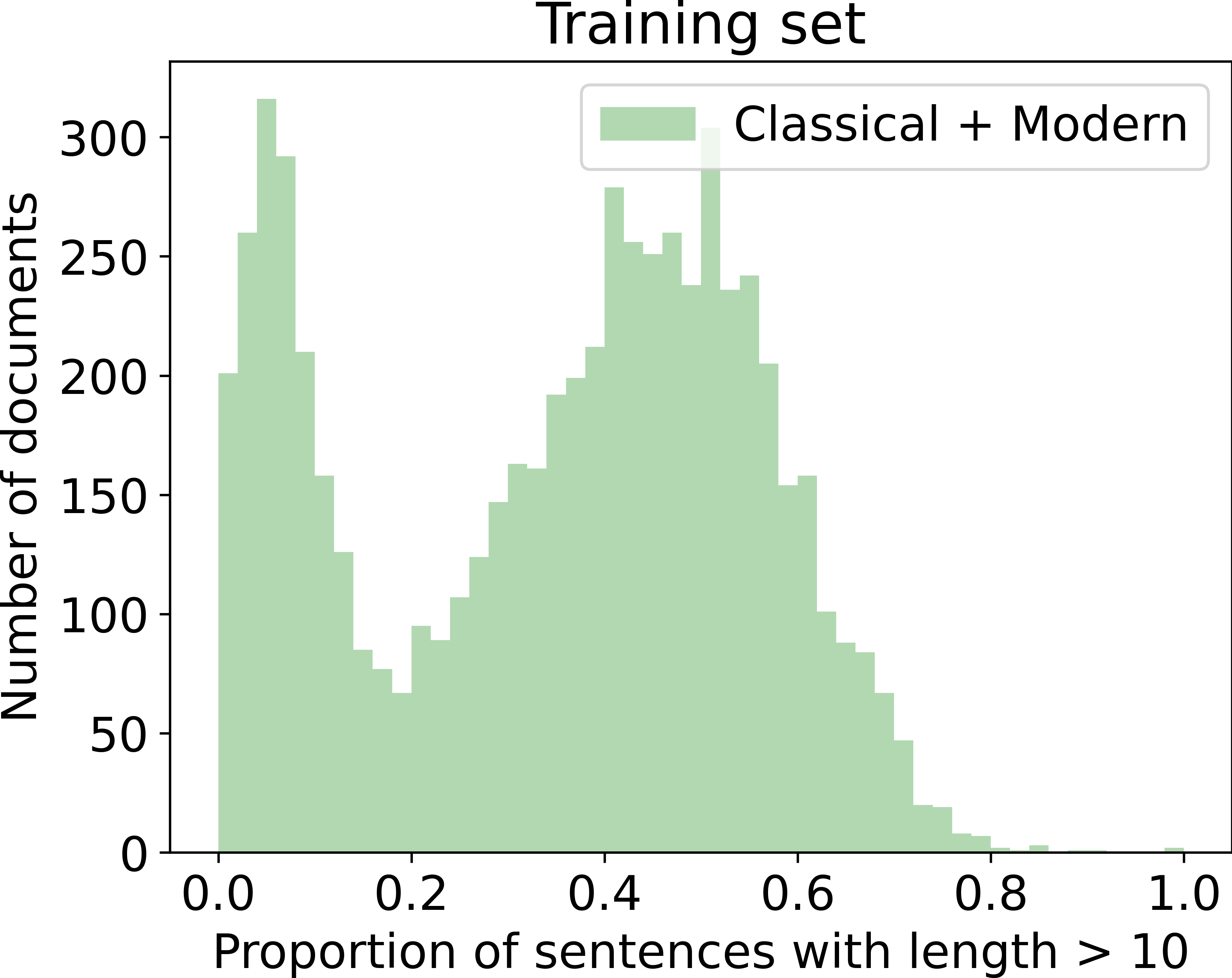}
  \caption{\xss{The histogram of $p_{(s>10)}$ over the training set. Although the writing style of these documents are not annotated, we could still observe two modalities with a separation boundary at $0.2$, which we will use as a rough criterion to distinguish modern and classical documents in the training set.}}
    \label{fig:training}
    \end{minipage}
\end{figure}

\xss{In addition to human annotation, we found that sentences in classical documents are usually shorter than in modern documents, which can be used as a simple criterion to categorize writing style. In detail, we split each document into sentences, compute the proportion of sentences with a length greater than 10, and denote it as $p_{(s>10)}$. 
We plot the histogram of $p_{(s>10)}$ in Figure~\ref{fig.dev+test} and~\ref{fig:training}. 
We remark that in validation and test sets, 98\% of the classical documents have $p_{(s>10)}<0.2$ while 96\% of the modern documents have $p_{(s>10)}\ge 0.2$. This suggests an approximate yet effective categorization criterion, i.e., $p_{(s>10)}<0.2$, for classifying document style over the training set.
}

\subsection{Question Type}
\label{sec:question_type}
To perform fine-grained analysis of the questions in {\name}, we conduct human annotations for a sampled batch of 300 questions from the test set. The label of each question is on the consensus of 3 annotators. 
The questions are categorized into 5 different categories:

\textbf{Matching} questions ask about a fact that has been explicitly described in the document. The correct answer can be directly obtained from a short span or a single sentence from the document. Note that different options can refer to different spans. 

\textbf{Semantic} questions ask about the semantic meanings of words or characters in a sentence, including antonym, synonymy, rhetoric and word segmentation. Q1 in Table~\ref{tab:examples} belongs to this category. We find that semantic questions are usually associated with classical Chinese documents.

\textbf{Summary} questions require the readers to understand all the facts stated throughout the entire document in order to choose a desired option, which presents a correct or incorrect fact summary.

\textbf{Reasoning} questions require the reader to perform non-trivial reasoning to infer a conclusion  not explicitly stated in the document. A reasoning question in {\name} often requires the reader to strong background knowledge and common sense. Q3 in Table~\ref{tab:examples} and Q1 in Table~\ref{tab:reasoning_questions} belong to this category.

\textbf{Sentiment} questions ask about the implicit sentiment that the author expressed in the document. Sentiment questions in {\name} typically require knowledge of imagery, symbolism and even the author's sociopolitical perspective.  Q2 in Table~\ref{tab:examples} belongs to this category.

\begin{table}[b]
    \caption{Distribution of question types in {\name} and {\Cs}.}
    \label{tab:reasoning_type}
    \centering
    \begin{tabular}{c|c c c c c}
    \hline
      Type   & {\name} & $\mathrm{C^3_M}$ & $\mathrm{C^3_D}$ & $\mathrm{C^3}$ \\
    \hline
        Matching & 0.33\% & 47.0\% &48.6\% & 47.8\% \\
        Semantic & 20.0\% & 1.7\% &2.3\% & 2.0\% \\
        Summary & 38.3\% & 15.3\%& 4.3\% & 9.8\% \\
        Reasoning & 28.3\% & 29.3\% & 36.7\% & 33.0\% \\
        Sentiment & 13.0\% & 6.7\% & 8.0 \% & 7.4\% \\
    \hline
    \end{tabular}
\end{table}

The annotations are summarized in Table~\ref{tab:reasoning_type}. We can observe that {\name} has very few matching questions, which indicates that most {\name} questions require non-trivial comprehension of the documents. 

As a comparison, we also sampled a total of 600 questions from {\Cs} dataset, another exam-question-based MRC dataset, with 300 from {\CsM} and 300 from {\CsD} respectively, and annotated the sampled questions with the same standard and annotation process.
The statistics are summarized in Table~\ref{tab:reasoning_type}. We can observe that {\Cs} contains a large portion of matching questions and much fewer summary and sentiment questions. Despite the fact that {\name} and {\Cs} has about the same percentage of reasoning questions, we remark that reasoning questions in {\name} are significantly harder than those in {\Cs}. This is not only because the documents in {\name} are longer (so that fact extraction will be harder) but also because the reasoning questions in {\name} typically require reasoning over a combination of document-level facts and background knowledge of both Chinese history and culture. To better illustrate the difference between {\name} and {\Cs}, we select two example reasoning questions  in Table~\ref{tab:reasoning_questions}, with one from {\Cs} and one from {\name} respective. 


\begin{table}[ht]
    \caption{Examples of reasoning questions from NCR (top) and {\Cs} (bottom) with Chinese (left) and English translation (right). We defer the NCR document to Table~2 in Appendix~C. $\star$ denotes the correct option. In order to correctly answer Q1 from {\name}, the reader not only needs to comprehend the texts in D1 describing the scene where Jun-tu meets the author's mother but also needs to understand the cultural meaning of  (madam).}
    \label{tab:reasoning_questions}
    \centering
    \small
    \begin{tabular}{m{0.8cm} | m{5cm} m{7cm}}
    \toprule
        \multirow{13}{*}{NCR}
        & D1: \begin{CJK}{UTF8}{gkai} 故乡（节选）鲁迅 \end{CJK}& D1: \emph{My old home} (excerpt) \textbf{Lu Xun}\\
        \cline{2-3}
        &Q1: \begin{CJK}{UTF8}{gkai}选出与本选段中中年闰土形象分析不恰当的一项\end{CJK} & Q1: Select the incorrect analysis of middle-aged Jun-tu's image. \\
        &\begin{CJK}{UTF8}{gkai}{A. 他称“我”母亲为“老太太”，表现了他有意讨好“我”母亲。}$\star$\end{CJK} & \textbf{A.He called my mother the ``madam'', which shows his intention to please my mother.} $\star$ \\
        &\begin{CJK}{UTF8}{gkai}B.他称自己少年时的好友为“老爷”，说明了他受封建等级观念影响很深。 \end{CJK} & B. He called his former good friend ``master'', which shows that he was deeply influenced by the feudal concept of hierarchy.  \\
        &\begin{CJK}{UTF8}{gkai}C.从他的对话中可以看出他的生活景况非常不好，他是当时下层人民形象的缩影。\end{CJK} &C. From his dialogue, we can see that the situation of his life is very bad, and he is the epitome of the image of the lower class at that time.\\
        &\begin{CJK}{UTF8}{gkai}D.宏儿和水生就像当年的“我”和闰土一样，彼此之间没有隔阂。\end{CJK} &D. Hung-eth and the Shui-sheng are just like me and Jun-tu at that time, they are not estranged from each other.\\
    \bottomrule
    \toprule
    \multirow{4}{*}{\Cs}
    &D2: \begin{CJK}{UTF8}{gkai} 男：还能不能再便宜点儿？” 女：“已经给您打五折了，先生！” \end{CJK}& D2. Man: Can you make it a little cheaper? " Woman: "I've given you a 50\% discount, sir!" \\
    \cline{2-3}
    &Q2: \begin{CJK}{UTF8}{gkai} 他们最可能是什么关系？ \end{CJK} & Q2. What is the most likely relationship between them? \\
    &\begin{tabular}{m{5.3cm}}
        \begin{CJK}{UTF8}{gkai} A.夫妻\quad C.老师和学生\end{CJK} \\ 
        \begin{CJK}{UTF8}{gkai} B.同事 \end{CJK}\quad  \begin{CJK}{UTF8}{gkai} \textsf{D.售货员和顾客} $\star$ \end{CJK}
    \end{tabular}
 & 
    \begin{tabular}{m{8cm}}
    A. Husband and wife \quad C. Teacher and student  \\ 
    B. Colleagues  \quad    \textbf{D. Salesperson and customer} $\star$
    \end{tabular}\\
    \bottomrule
    \end{tabular}
\end{table}


\section{Experiment}
\label{experiment setting}

In this section, we conduct quantitative study as well as human evaluation on our {\name} dataset. 

\subsection{Baseline Methods}
\paragraph{Trivial Baselines:} We consider random guess and deterministic choice as trivial baselines. Deterministic choice always selects the same option ID.

\paragraph{Fine-Tuning of Pretrained Model: } We utilize the MRC model architecture from the BERT paper~\cite{BERT} and perform fine-tuning with {\name}. We consider the fine-tuned performance of 7 popular Chinese pretrained models, including BERT-Chinese~\cite{BERT}, ERNIE~\cite{ERNIE}, BERT-wwm, BERT-wwm-ext, RoBERTa-large-Chinese~\cite{WWM}, MacBERT-base and MacBERT-large~\cite{MacBERT}.
We also investigate the effectiveness of data augmentation by additionally collecting 6000 documents and 13K exam questions for China's primary-school Chinese course. We combine these primary-school exam questions and the {\name} training data as an augmented dataset to further boost the final performance. 
All the model and training details can be found in appendix~B. 

\paragraph{Competition:}
To examine the limit of current MRC methods, we organized a 3-month-long online competition using {\name} with training and validation set released. Participants are allowed to use any open-access pretrained model or any open-access \emph{unlabeled} data. Use of any external MRC \emph{supervision} is forbidden, since a portion of the test questions are possibly accessible online. This aims to prevent human annotations overlapping with our held-out data for a fair competition. There are a total of 141 participating teams and the best submission model with the highest test accuracy is taken as the \emph{competition} model. The team is from \xss{an} industry lab. They first pre-trained an XLNet-based model~\cite{XLNET} on a company-collected large corpus\footnote{Unfortunately, the company disagreed to release their internal pretraining data but the final trained model will be released at our project website.}. For each question, they use an information retrieval tool Okapi BM25~\cite{robertson1976relevance, robertson2009probabilistic} to extract the most relevant parts from the document and then run this pre-trained model for answer selection based on the extracted texts. 



\paragraph{Human Evaluation:}
We randomly sample 50 documents with 120 questions from the annotated subset of test data in {\name} and send these questions to 30 sophomore college students. All the students are native Chinese speakers majored in computer science, who have not \xss{taken} any Chinese course during the recent 2 years after college admission. Therefore, we believe they have reasonable reading comprehension capabilities close to typical China's high school students.  Each question is completed by at least 3 students to get an accurate performance estimation.


\subsection{Results}
\label{sec:result}

\subsubsection{Overall Performance}
\label{sec:overall_perf}
The performance of different baseline methods as well as human volunteers are summarized in Table~\ref{tab:results}. Pretrained models are substantially better than trivial baselines. Particularly, the MacBERT-large model produces the highest fine-tuning test accuracy of \xss{0.4780} while the use of external data augmentation further boost the test performance to \xss{0.5021}, which suggest the effectiveness of data augmentation. 
The best model comes from the participants of the online competition. The best competition model achieves a test accuracy of \xss{0.5985}, which is much higher than the best fine-tuning model that authors obtained. 
However, the human volunteers achieves an average accuracy of \xss{0.7917}, which results in a 20\% performance margin over the best baseline MRC model. \xss{Human performance is averaged per question over three annotators. To measure the inter-agreement, we calculate the agreement ratios. 60.83\% of the questions have the same answer from all the 3 students, and 96.67\% of the questions have the same answers from at least 2 students.}
\begin{table}[ht]
    \caption{Validation and test accuracy of different MRC methods on {\name}. \textbf{*}~Human evaluation is only conducted over a subset of test questions.}
    \label{tab:results}
    \centering
    \begin{tabular}{c c c}
    \hline
      Method   & Val. & Test \\
    \hline
        Random Guess & 0.2505 & 0.2511 \\
        Deterministic Choice & 0.2951 & 0.2613\\
    \hline
       BERT-Chinese & 0.3930 & 0.3946 \\
       ERNIE & 0.4445 & 0.4252\\
       BERT-wwm & 0.4310 & 0.4272\\
       BERT-wwm-ext & 0.4814 &0.4451  \\
       MacBERT&  0.4736&0.4597  \\
       RoBERTa-large-Chinese & 0.4666 & 0.4642\\
       MacBERT-large& 0.5051 & 0.4780 \\
       MacBERT-large \footnotesize{(data aug.)} &\textbf{ 0.5199} & \textbf{0.5021} \\
    \hline
       Competition & 0.5831 & 0.5985 \\
    \hline
        Human performance& N/A  & 0.7917* \\
      \hline
    \end{tabular}
\end{table}
\begin{table}[ht]
    \caption{Test accuracy of human and AI w.r.t. different document writing styles. FT is the best model finetuned by ourselves and CMP is the best competition model.}
    \label{tab:article_type_result}
    \centering
    \begin{tabular}{c|ccc}
    \hline
        Document Style & Human & FT & CMP   \\
    \hline
        Modern &  0.7489  & 0.5257 & 0.6151\\
        Classical \scriptsize (w/o poetry) & 0.8632  & 0.4502 & 0.5671\\
        Poetry only & 0.9167  & 0.3462 & 0.4179 \\
    \hline
    \end{tabular}
\end{table}
\begin{table}[ht]
    \caption{\xss{Test accuracy of MacBERT-large fine-tuned on complete and filtered training set.}}
    \label{tab:modern_only}
    \centering
    \setlength{\belowcaptionskip}{-0.3cm}
    \begin{tabular}{c|ccc}
    \hline
        Training set & Modern & Classical \scriptsize (w/o poetry) & poetry  \\
    \hline
        Complete &  0.5039  & 0.3871 & 0.3582  \\
        Filtered & 0.5060  & 0.2970 & 0.2836  \\
    \hline
    \end{tabular}
\end{table}
\begin{table}[ht]
    \caption{Test accuracy of human and AI w.r.t. different document lengths in both classical and modern Chinese. Human data are only presented when at least 5 documents can be collected from the annotated subset.}
    \label{tab:document_len_result}
    \centering
    \begin{tabular}{c|cccc}
    \hline
    \multicolumn{5}{c}{Classical} \\
    \hline
        Len. & \footnotesize[0,100]&\footnotesize(100, 300] & \footnotesize(300, 600]&>\footnotesize600\\
    \hline
        FT &  0.3014 & 0.5505 & 0.4162 & 0.4203 \\
        CMP & 0.2192 & 0.5046 & 0.6069 & 0.6116 \\
        Human & N/A & 0.8333 & 0.7333 & 0.8958 \\
    \hline
    \multicolumn{5}{c}{Modern} \\
    \hline
        Len. & \footnotesize[0,300]& \footnotesize (300, 600] &
        \footnotesize (600, 1200] & \footnotesize > 1200 \\
    \hline
        FT & 0.5714 & 0.4203 & 0.5220 & 0.5486 \\
        CMP & 0.3809 & 0.5652 & 0.6373 & 0.5973 \\
        Human & N/A & N/A & 0.7879 & 0.6970 \\
    \bottomrule
    \end{tabular}
\end{table}

\subsubsection{Fine-Grained Analysis}
We measure the performance of the MRC models, i.e., the best fine-tuned model (\textbf{\emph{FT}}) and the competition model (\textbf{\emph{CMP}}), and human on the test questions w.r.t. different factors, including writing style, document length and question type. 
We remark that the model accuracy are measured over the entire test set except the study on question type, which are over the annotated subset only.
\paragraph{Writing style:}
Table~\ref{tab:article_type_result} illustrates the performance of MRC models and human on different document styles. The performance of AI significantly drops on classical Chinese documents, particularly on poetry. By contrast, we observe an opposite phenomena for humans, who perform the best on poetry, the most abstract form of classical Chinese literature. \xss{We remark that in China's Chinese exams, questions on modern and classical texts may often have different examination focuses.
Questions on classical Chinese are more biased towards understanding the meaning of characters, words, and sentences (see Table~\ref{tab:short_examples}, Q1), which may not be intrinsically difficult for native Chinese students who have been well trained. While modern Chinese questions are often more general and require in-depth understanding of the entire document (see Table~\ref{tab:short_examples}, Q3), which can be challenging for humans. 
This is because Chinese documents are much longer and reading and remembering facts under long documents can easily make a human distracted.
}

\xss{We also investigate whether a AI model purely trained on modern Chinese can directly transfer to classical texts. 
Since we do not have ground truth annotations on training set, we following the filtering process in section~\ref{article_style} as a rough categorization, which yields 4507 documents.
We fine-tune the MacBERT-large on this filtered training set (without data augmentation) and show the testing results  in Table~\ref{tab:modern_only}.
We can observe that the performance on classical documents drops significantly with classical documents filtered out, while the performance on modern documents remains unchanged.
Hence, we argue that classical Chinese training data can be critical.
}
This also suggests an important direction for improving Chinese pre-trained models.


\paragraph{Document Length:}
Table.~\ref{tab:document_len_result} summarizes the performance of human and AI on documents with various lengths. 
For classical Chinese documents, the most challenging documents are those shortest ones, which are most likely poetry.
For modern articles, human performance drops for particularly long articles. For AI, the hierarchically-structured competition model performs the worst on relatively short articles while the fine-tuning model has the most difficulties in documents with a moderate length, i.e., from 300 to 600 characters. This suggests possible enhancements on model architecture.
Moreover, although the performance gap between human and machine becomes less significant on particularly long documents, the accuracy of AI systems remains unsatisfying in general.
We also want to remark that even those relative short documents in {\name} are substantially longer and more sophisticated than existing datasets of the similar question types like \Cs.



\begin{table}[ht]
    \caption{Test accuracy of human and AI w.r.t. different question types. We ignore matching questions since they are too infrequent.}
    \label{tab:reasoning_type_result}
    \centering
    \setlength{\belowcaptionskip}{-0.3cm}
    \begin{tabular}{c|c|c|c}
    \hline
        Question Type & Human & FT & CMP  \\
    \hline
        Semantic &  0.9047  & 0.5000 & 0.5833  \\
        Summary & 0.7976  & 0.5431 & 0.5603  \\
        Reasoning & 0.7179 & 0.6000 & 0.6588 \\
        Sentiment & 0.6333 & 0.5641 & 0.5128 \\
    \hline
    \end{tabular}
\end{table}
\paragraph{Question Type:}
We also compare the performance of human and AI on different question types in Table.~\ref{tab:reasoning_type_result}. To our surprise, sentiment questions, which are the most challenging for humans, yield the smallest performance margin. While the largest gap is on semantic questions, which we believe the easiest for human. This indicates that a pretrained model is capable of capturing high-level sentiment information but still lacks word/character-level reasoning abilities. In addition, we also observed that the hierarchical competition model performs much worse than the fine-tuned model on sentiment questions, which suggests that running a retrieval model first may result in a loss of document-level global information which can be critical for sentiment analysis.
This raises an open challenge for building more effective hierarchical models for processing long texts.

\section{Conclusion}

We present a novel Chinese MRC dataset, \emph{Native Chinese Reader (\name)}, towards building \emph{native-level} Chinese MRC models. Experiments on {\name} indicate a significant gap between current MRC methods and human performance, which suggests great opportunities for future research, and, hopefully, pushes the frontier of Chinese natural language understanding.

\textbf{Remark:} Our dataset primarily consists of open-access exam questions or generated ones with teacher permission. All the documents are all public teaching materials. The released models are permitted by the online competition participants. Annotations and human evaluation results are completed by PhD students and interns that are all paid according to our institute regulation. Hence, we believe that our project will not lead to any legal or ethical issues.

\subsection*{Acknowledgements}
Yi Wu is supported by 2030 Innovation Megaprojects of China (Programme on New Generation Artificial Intelligence) Grant No. 2021AAA0150000. We would also like to thank the anonymous reviewers for their insightful feedbacks.
\bibliography{anthology,custom}
\bibliographystyle{plain}

\newpage
\appendix
{\LARGE $$\makebox{\textbf{Supplementary Materials}}$$}
\section{Project Statement}
\textbf{Dataset: } All the materials are collected from online education materials or generated by qualified high-school teachers. All the documents are manually verified by the authors so that they all follow the China's Chinese education standard, which does not contain any sensitive or inappropriate content for high school students. Language of the dataset is written by Mandarin Chinese in the simplified script per China's education system regulation, including the classical Chinese literature.
A detailed dataset instruction as well as a full dataset statement can be found at our project website, \url{https://sites.google.com/view/native-chinese-reader/}.

\textbf{Accessibility: } The dataset, related models as well as our human annotations are all kept at Google drive with links at our project website. All of our authors promise to keep the best efforts to keep them accessible. Our project website also has a detailed introduction to the online competition we organized. We include the requirement that all the participating teams should release their model in our competition agreement, so the release of models are permitted by all the participants. 

\textbf{Human Accuracy: } Note that each problem requires non-trivial reasoning and reading a long article. To ensure the students treat the problems seriously, we leave this task as a bonus homework in our deep learning course for second-year college students at Tsinghua university. We also make sure that each student will be assigned no more than 10 articles for a reasonable workload. We also make sure each problem is at least answered by 3 students to get an accurate estimate. 

\textbf{Licence:} This dataset is released under the CC BY-SA 4.0 license for general research purpose.
\section{Model Details}
\label{sec:model_details}

Our our code can be found at our project website and are released under the MIT license.

The baseline model of this article is based on pre-trained BERT,  with some modifications made to the input layer and output layer. 

The input of this task consists of three parts, which are articles, questions, and options. There are 2-4 options for each question, we fill the questions into 4 options. For convenience, we denote the document as D, the question as Q, and the answer options as $A_1$, $A_2$, $A_3$, $A_4$. For the \textit{i}-th option, we construct an input sequence as [CLS] D [SEP] Q [SEP] $A_i$ [SEP]. \xss{We truncated  the documents by dropping the end part to ensure that the input sequences are not longer than the maximum positions of the pre-trained models.}

We fine-tuned several commonly used Chinese pre-trained models as baselines, including Google's BERT-chinese, Baidu's ERNIE, HFL's BERT-wwm, BERT-wwm-ext, and MacBERT ~\cite{BERT, WWM, MacBERT}. These 5 base models have a similar model structure with 12-layer of Transformer Encoder~\cite{Transformer} and 12 attention heads, the hidden size is 768. We also tried two large models, Roberta-large and MacBERT-large, both of which have 24 layers and 16 attention heads, hidden size is 1024.

We compared the results of the 7 baseline models in section~4.2.
The BERT-based model will generate a 768-dimensional hidden state for each token of the input sequence. We take the output vector of [CLS] token and map it to a \emph{one-dimensional} logit through a trainable vector. The logit means to what extent $A_i$ may be the correct option. In the same way, we can get the logit of the other 3 options, and employ Softmax to compute the probability of the 4 options.
The training objective is to minimize the four-class cross-entropy. In the inferring stage, we just select the option with the highest probability.

All baseline models are trained on 8 Tesla V100 GPU with 32G memory. We set the same hyper-parameters for all base models and large models, separately. For base models, we set epoch as 10, batch size as 64, learning rate as 5e-6. For large models, we set epoch as 10, batch size as 32, learning rate as 2e-6. Regarding data augmentation, we first use the primary school dataset for 5 epoch and then turn to NCR dataset for 5 epochs.  All the codes are implemented based on Hugging Face transformers~\cite{HuggingFace}. It takes about 5 minutes to run one epoch for base model and 20 minutes for large model. 

The best competition model applied a similar structure. However, they cut the document into several chunks and utilize an information retrieval tool Okapi BM25 to extract the most relevant chunk according to the question as to the actual passage, which reduce the super long document to a reasonable length. The document segment, question and the options are then input to XLNet-based classifier to predict the correct answer.
Compared to our baseline model, their IR tool can effectively keep the most informative segment of the document while reducing to a reasonable sequence length. 

\section{Surface pattern in options}
\begin{CJK}{UTF8}{gkai}
\xss{We investigate some special patterns and predict the answer based on these patterns. In Table~\ref{tab:pattern}, we focus on several absolute quantifiers including ``只能'' (``can only''), ``必然'', ``必定'', ``必须'', (``must''), ``只可能'', (``may only''), ``绝对'' (``absolute''). For each pattern, We keep the questions where only one option contains this pattern or only one option doesn't contain this pattern and choose the special option as the predicted answer. The results are aggregated from the whole dataset with a total of 20477 questions. ``combination’’ represents a combination pattern that including all the aforementioned patterns. We can observe that there are not many questions meeting the requirements, and the accuracy is indeed higher than random but also still very low.}
\begin{table}[ht]
    \centering
    \caption{\xss{Predict the answer based on special patterns. The results are aggregated from the whole dataset with a total of 20477 questions. ``combination’’ represents a combination pattern that including all the aforementioned patterns.}}
    \label{tab:pattern}
    \begin{tabular}{c|ccccccc}
    \toprule
        Pattern & 只能 & 必然 & 必定 & 必须 & 只可能 & 绝对 & combination \\
    \midrule
        \# Question & 291 & 308 & 52 & 658 & 5 & 101 & 1296 \\ 
        Accuracy & 0.3470 & 0.2825 & 0.3077 & 0.2903 & 0 & 0.2673 & 0.2994 \\
    \bottomrule
    \end{tabular}
\end{table}
\end{CJK}

\section{Additional Examples}
\label{sec:additional_examples}
Limited by space, we show some additional examples in this section. The example in Table~\ref{tab:classical_chinese} is a classical Chinese with a question of sentiment. It described a dialogue between the author and his friend, which reflects the author's inner contradictions and complex feelings. The author quoted predecessors' (\begin{CJK}{UTF8}{gkai}曹操 Cao Cao\end{CJK}) verses and the allusions to \textbf{the Battle of Chibi} (\begin{CJK}{UTF8}{gkai}赤壁之战\end{CJK}) to express his feelings. The question requires the readers to analyze the document from different perspectives, including the author's sentiment, writing techniques, and rhyme.

In Table~\ref{tab:hometown}, we present the document of question in Table~6 of the main paper with its English translation, which is a excerpt from Lu Xun's famous article \emph{My old home} (sometimes it is translated to \emph{Hometown})

\begin{table*}[!htbp]
    \footnotesize

    \centering
        \begin{tabular}{m{5cm}  m{8cm}}
        \hline
        \begin{CJK}{UTF8}{gkai}
        苏子愀然，正襟危坐，而问客曰：“何为其然也？”客曰：“月明星稀，乌鹊南飞，此非曹孟德之诗乎？西望夏口，东望武昌，山川相缪，郁乎苍苍，此非孟德之困于周郎者乎？方其破荆州，下江陵，顺流而东也，舳舻千里，旌旗蔽空，酾酒临江，横槊赋诗，固一世之雄也，而今安在哉？况吾与子渔樵于江渚之上，侣鱼虾而友糜鹿，驾一叶之扁舟，举匏樽以相属。寄蜉蝣与天地，渺沧海之一粟。哀吾生之须臾，羡长江之无穷。挟飞仙以遨游，抱明月而长终。知不可乎骤得，托遗响于悲风。”苏子曰：“客亦知夫水与月乎？逝者如斯，而未尝往也；盈虚者如彼，而卒莫消长也。盖将自其变者而观之，而天地曾不能一瞬；自其不变者而观之，则物与我皆无尽也，而又何羡乎？且夫天地之间，物各有主，苟非吾之所有，虽一毫而莫取。惟江上之清风，与山间之明月，耳得之而为声，目遇之而成色。取之无禁，用之不竭，是造物者之无尽藏也，而吾与子之所共适。 
        \end{CJK} &
        Sad at heart, I sat up straight to ask my friend why the music was so mournful. He replied, “Didn't Cao Cao describe a scene like this in his poem: 'The moon is bright, the stars are scattered, the crows fly south...?' And isn't this the place where he was defeated by Zhou Yu? See how the mountains and streams intertwine, and how darkly imposing they are with Xiakou to the west and Wuchang to the east. When Cao Cao took Jingzhou by storm and conquered Jiangling, then advanced eastward along the river, his battleships stretched for a thousand li, his armies' pennons and banners filled the sky. When he offered a libation of wine on the river and lance in hand chanted his poem, he was the hero of his times. But where is he now? We are mere fishermen and woodcutters, keeping company with fish and prawnsand befriending deer. We sail our skiff, frail as a leaf, and toast each other by drinking wine from a gourd. We are nothing but insects who live in this world but one day, mere specks of grain in the vastness of the ocean. I am grieved because our life is so transient, and envy the mighty river which flows on forever. I long to clasp winged fairies and roam freely, or to embrace the bright moon for all eternity. But knowing that this cannot be attained at once, I give vent to my feelings in these notes which pass with the sad breeze. ” Then I asked him, “Have you considered the water and the moon? Water flows away but is never lost; the moon waxes and wanes, but neither increases nor diminished. If you look at its changing aspect, the universe passes in the twinkling of an eye; but if you look at its changeless aspect, all creatures including ourselves are imperishable. What reason have you to envy other things? Besides, everything in this universe has its owner; and if it does not belong to me not a tiny speck can I take. The sole exceptions are the cool breeze on the river, the bright moon over the hills. These serve as music to our ears, as colour to our eyes; these we can take freely and enjoy forever; these are inexhaustible treasures supplied by the Creator, and things in which we can delight together.
        \\
         \hline
        \textbf{Q} \begin{CJK}{UTF8}{gkai}
        下列对文段的理解和分析不正确的一项是
        \end{CJK}
         &\textbf{Q} Choose one from the following options which is incorrect understanding and analysis of the document.\\
        A. \begin{CJK}{UTF8}{gkai}
        主客对话的内容实际上是作者内心思想矛盾和感情复杂的反映，运用主客问答的方式，可以使行文波澜起伏，摇曳多姿，作者的思想感情也因此充分展现。
        \end{CJK} &
        A. The content of the dialogue between the author and his friend is actually a reflection of the author’s inner contradictions and complex feelings. The use of question-and-answer methods can make the writing ups and downs, swaying, and fully demonstrate the author’s thoughts and feelings.\\
        B. \begin{CJK}{UTF8}{gkai}
        作者通过多组对比，揭示出“悲”的原因。文中既写了曹孟德一世之雄的兴亡之悲，也写了由宇宙无穷与人生短暂的对比所生之悲，还写了现实与理想的矛盾所生之悲。
        \end{CJK} & 
        B. The author reveals the cause of "sorrow" through multiple groups of comparisons. The article not only writes the tragedy of the rise and fall of Cao Cao, but also the tragedy born from the contrast between the infinity of the universe and the short-lived life, and the tragedy born from the contradiction between reality and ideal.\\
        C. \begin{CJK}{UTF8}{gkai}
        作者以江水明月作比，说明世间万物和人生既有变的一面，也有不变的一面，阐述了自然万物变化与永恒的哲理，表现出作者无奈消极的人生态度。$\star$
        \end{CJK} & 
        \textbf{C. The author uses the river, water and the moon as a comparison to illustrate that everything in the world and life has both a changeable side and an unchanging side. He expounds the change and eternal philosophy of all things in nature, showing the author's helpless and passive attitude towards life.} $\star$\\
        D. \begin{CJK}{UTF8}{gkai}
        文段句式骈散结合，结构、句法、韵律都相对自由。大量对偶句的使用使文章参差错落，整齐简约，极富声韵之美。
        \end{CJK} &
        D. The paragraphs and sentences are combined with parallel and prose, and the structure, syntax, and rhythm are relatively free. The use of a large number of antithetical sentences makes the article jumbled, neat and simple, and full of beauty of rhyme and rhyme.\\
    \hline
    \end{tabular}
    \caption{An example of document in Classical Chinese with questions (left) and English translation (right), $\star$ is the correct option. And this question is an example of sentiment}
    \label{tab:classical_chinese}
\end{table*}

\begin{table*}[ht]
    \centering
    \begin{tabular}{m{13cm}}
    \toprule
    \begin{CJK}{UTF8}{gkai}
    故乡（节选）我这时很兴奋，但不知道怎么说才好，只是说：“阿！闰土哥，——你来了？……” 我接着便有许多话，想要连珠一般涌出：角鸡，跳鱼儿，贝壳，猹……但又总觉得被什么挡着似的，单在脑里面回旋，吐不出口外去。 他站住了，脸上现出欢喜和凄凉的神情；动着嘴唇，却没有作声。他的态度终于恭敬起来了，分明的叫道：“老爷！……” 我似乎打了一个寒噤；我就知道，我们之间已经隔了一层可悲的厚障壁了。我也说不出话。 他回过头去说，“水生，给老爷磕头。”便拖出躲在背后的孩子来，这正是一个廿年前的闰土，只是黄瘦些，颈子上没有银圈罢了。“这是第五个孩子，没有见过世面，躲躲闪闪……” 母亲和宏儿下楼来了，他们大约也听到了声音。 “老太太。信是早收到了。我实在喜欢的了不得，知道老爷回来……”闰土说。 “阿，你怎的这样客气起来。你们先前不是哥弟称呼么？还是照旧：迅哥儿。”母亲高兴的说。 “阿呀，老太太真是……这成什么规矩。那时是孩子，不懂事……”闰土说着，又叫水生上来打拱，那孩子却害羞，紧紧的只贴在他背后。 “他就是水生？第五个？都是生人，怕生也难怪的；还是宏儿和他去走走。”母亲说。 宏儿听得这话，便来招水生，水生却松松爽爽同他一路出去了。母亲叫闰土坐，他迟疑了一回，终于就了坐，将长烟管靠在桌旁。
    \end{CJK} \\
    \midrule
    \textbf{Translation:} \\
    \emph{My old home} (excerpt) \textbf{Lu xun} Delighted as I was, I did not know how to express myself, and could only say: "Oh! Jun-tu—so it's you? . . ."
    After this there were so many things I wanted to talk about, they should have poured out like a string of beads: woodcocks, jumping fish, shells, zha. . . . But I was tongue-tied, unable to put all I was thinking into words.
    He stood there, mixed joy and sadness showing on his face. His lips moved, but not a sound did he utter. Finally, assuming a respectful attitude, he said clearly:
    "Master! . . ."
    I felt a shiver run through me; for I knew then what a lamentably thick wall had grown up between us. Yet I could not say anything.
    He turned his head to call:
    "Shui-sheng, bow to the master." Then he pulled forward a boy who had been hiding behind his back, and this was just the Jun-tu of twenty years before, only a little paler and thinner, and he had no silver necklet.
    "This is my fifth," he said. "He's not used to company, so he's shy and awkward."
    Mother came downstairs with Hung-erh, probably after hearing our voices.
    "I got your letter some time ago, madam," said Jun-tu. "I was really so pleased to know the master was coming back. . . ."
    "Now, why are you so polite? Weren't you playmates together in the past?" said mother gaily. "You had better still call him Brother Hsun as before."
    "Oh, you are really too. . . . What bad manners that would be. I was a child then and didn't understand." As he was speaking Jun-tu motioned Shui-sheng to come and bow, but the child was shy, and stood stock-still behind his father.
    "So he is Shui-sheng? Your fifth?" asked mother. "We are all strangers, you can't blame him for feeling shy. Hung-erh had better take him Out to play."
    
    When Hung-eth heard this he went over to Shui-sheng, and Shui-sheng went out with him, entirely at his ease. Mother asked Jun-tu to sir down, and after a little hesitation he did so; leaning his long pipe against the table.\\
    \bottomrule

    \end{tabular}

    \caption{The document (top) of example in Table~6 and its English translation. (bottom)}
    \label{tab:hometown}
\end{table*}



\end{document}


\maketitle


\appendix

\section{Project Statement}
\textbf{Dataset: } All the materials are collected from online education materials or generated by qualified high-school teachers. All the documents are manually verified by the authors so that they all follow the China's Chinese education standard, which does not contain any sensitive or inappropriate content for high school students. Language of the dataset is written by Mandarin Chinese in the simplified script per China's education system regulation, including the classical Chinese literature.

\textbf{Accessibility: } The dataset, related models as well as our human annotations are all kept at Google drive with links at our project website. All of our authors promise to keep the best efforts to keep them accessible. Our project website also has a detailed introduction to the online competition we organized. We include the requirement that all the participating teams should release their model in our competition agreement, so the release of models are permitted by all the participants. 

\textbf{Human Accuracy: } Note that each problem requires non-trivial reasoning and reading a long article. To ensure the students treat the problems seriously, we leave this task as a bonus homework in our deep learning course for second-year college students at Tsinghua university. We also make sure that each student will be assigned no more than 10 articles for a reasonable workload. We also make sure each problem is at least answered by 3 students to get an accurate estimate. 

\textbf{Licence:} This dataset is released under the CC BY-SA 4.0 license for general research purpose.

\clearpage
\section{Model Details}
\label{sec:model_details}

Our our code can be found at our project website and are released under the MIT license.

The baseline model of this article is based on pre-trained BERT,  with some modifications made to the input layer and output layer. 

The input of this task consists of three parts, which are articles, questions, and options. There are 2-4 options for each question, we fill the questions into 4 options. For convenience, we denote the document as D, the question as Q, and the answer options as $A_1$, $A_2$, $A_3$, $A_4$. For the \textit{i}-th option, we construct an input sequence as [CLS] D [SEP] Q [SEP] $A_i$ [SEP]. \xss{We truncated  the documents by dropping the end part to ensure that the input sequences are not longer than the maximum positions of the pre-trained models.}

We fine-tuned several commonly used Chinese pre-trained models as baselines, including Google's BERT-chinese, Baidu's ERNIE, HFL's BERT-wwm, BERT-wwm-ext, and MacBERT ~\cite{BERT, WWM, MacBERT}. These 5 base models have a similar model structure with 12-layer of Transformer Encoder~\cite{Transformer} and 12 attention heads, the hidden size is 768. We also tried two large models, Roberta-large and MacBERT-large, both of which have 24 layers and 16 attention heads, hidden size is 1024.

We compared the results of the 7 baseline models in section~4.2.
The BERT-based model will generate a 768-dimensional hidden state for each token of the input sequence. We take the output vector of [CLS] token and map it to a \emph{one-dimensional} logit through a trainable vector. The logit means to what extent $A_i$ may be the correct option. In the same way, we can get the logit of the other 3 options, and employ Softmax to compute the probability of the 4 options.
The training objective is to minimize the four-class cross-entropy. In the inferring stage, we just select the option with the highest probability.

All baseline models are trained on 8 Tesla V100 GPU with 32G memory. We set the same hyper-parameters for all base models and large models, separately. For base models, we set epoch as 10, batch size as 64, learning rate as 5e-6. For large models, we set epoch as 10, batch size as 32, learning rate as 2e-6. Regarding data augmentation, we first use the primary school dataset for 5 epoch and then turn to NCR dataset for 5 epochs.  All the codes are implemented based on Hugging Face transformers~\cite{HuggingFace}. It takes about 5 minutes to run one epoch for base model and 20 minutes for large model. 

The best competition model applied a similar structure. However, they cut the document into several chunks and utilize an information retrieval tool Okapi BM25 to extract the most relevant chunk according to the question as to the actual passage, which reduce the super long document to a reasonable length. The document segment, question and the options are then input to XLNet-based classifier to predict the correct answer.
Compared to our baseline model, their IR tool can effectively keep the most informative segment of the document while reducing to a reasonable sequence length. 

\section{Surface pattern in options}
\begin{CJK}{UTF8}{gkai}
\xss{We investigate some special patterns and predict the answer based on these patterns. In Table~\ref{tab:pattern}, we focus on several absolute quantifiers including ``只能'' (``can only''), ``必然'', ``必定'', ``必须'', (``must''), ``只可能'', (``may only''), ``绝对'' (``absolute''). For each pattern, We keep the questions where only one option contains this pattern or only one option doesn't contain this pattern and choose the special option as the predicted answer. The results are aggregated from the whole dataset with a total of 20477 questions. ``combination’’ represents a combination pattern that including all the aforementioned patterns. We can observe that there are not many questions meeting the requirements, and the accuracy is indeed higher than random but also still very low.}
\begin{table}[ht]
    \centering
    \caption{\xss{Predict the answer based on special patterns. The results are aggregated from the whole dataset with a total of 20477 questions. ``combination’’ represents a combination pattern that including all the aforementioned patterns.}}
    \label{tab:pattern}
    \begin{tabular}{c|ccccccc}
    \toprule
        Pattern & 只能 & 必然 & 必定 & 必须 & 只可能 & 绝对 & combination \\
    \midrule
        \# Question & 291 & 308 & 52 & 658 & 5 & 101 & 1296 \\ 
        Accuracy & 0.3470 & 0.2825 & 0.3077 & 0.2903 & 0 & 0.2673 & 0.2994 \\
    \bottomrule
    \end{tabular}
\end{table}
\end{CJK}

\section{Additional Examples}
\label{sec:additional_examples}
Limited by space, we show some additional examples in this section. The example in Table~\ref{tab:classical_chinese} is a classical Chinese with a question of sentiment. It described a dialogue between the author and his friend, which reflects the author's inner contradictions and complex feelings. The author quoted predecessors' (\begin{CJK}{UTF8}{gkai}曹操 Cao Cao\end{CJK}) verses and the allusions to \textbf{the Battle of Chibi} (\begin{CJK}{UTF8}{gkai}赤壁之战\end{CJK}) to express his feelings. The question requires the readers to analyze the document from different perspectives, including the author's sentiment, writing techniques, and rhyme.

In Table~\ref{tab:hometown}, we present the document of question in Table~6 of the main paper with its English translation, which is a excerpt from Lu Xun's famous article \emph{My old home} (sometimes it is translated to \emph{Hometown})

\begin{table*}[!htbp]
    \footnotesize

    \centering
        \begin{tabular}{m{5cm}  m{8cm}}
        \hline
        \begin{CJK}{UTF8}{gkai}
        苏子愀然，正襟危坐，而问客曰：“何为其然也？”客曰：“月明星稀，乌鹊南飞，此非曹孟德之诗乎？西望夏口，东望武昌，山川相缪，郁乎苍苍，此非孟德之困于周郎者乎？方其破荆州，下江陵，顺流而东也，舳舻千里，旌旗蔽空，酾酒临江，横槊赋诗，固一世之雄也，而今安在哉？况吾与子渔樵于江渚之上，侣鱼虾而友糜鹿，驾一叶之扁舟，举匏樽以相属。寄蜉蝣与天地，渺沧海之一粟。哀吾生之须臾，羡长江之无穷。挟飞仙以遨游，抱明月而长终。知不可乎骤得，托遗响于悲风。”苏子曰：“客亦知夫水与月乎？逝者如斯，而未尝往也；盈虚者如彼，而卒莫消长也。盖将自其变者而观之，而天地曾不能一瞬；自其不变者而观之，则物与我皆无尽也，而又何羡乎？且夫天地之间，物各有主，苟非吾之所有，虽一毫而莫取。惟江上之清风，与山间之明月，耳得之而为声，目遇之而成色。取之无禁，用之不竭，是造物者之无尽藏也，而吾与子之所共适。 
        \end{CJK} &
        Sad at heart, I sat up straight to ask my friend why the music was so mournful. He replied, “Didn't Cao Cao describe a scene like this in his poem: 'The moon is bright, the stars are scattered, the crows fly south...?' And isn't this the place where he was defeated by Zhou Yu? See how the mountains and streams intertwine, and how darkly imposing they are with Xiakou to the west and Wuchang to the east. When Cao Cao took Jingzhou by storm and conquered Jiangling, then advanced eastward along the river, his battleships stretched for a thousand li, his armies' pennons and banners filled the sky. When he offered a libation of wine on the river and lance in hand chanted his poem, he was the hero of his times. But where is he now? We are mere fishermen and woodcutters, keeping company with fish and prawnsand befriending deer. We sail our skiff, frail as a leaf, and toast each other by drinking wine from a gourd. We are nothing but insects who live in this world but one day, mere specks of grain in the vastness of the ocean. I am grieved because our life is so transient, and envy the mighty river which flows on forever. I long to clasp winged fairies and roam freely, or to embrace the bright moon for all eternity. But knowing that this cannot be attained at once, I give vent to my feelings in these notes which pass with the sad breeze. ” Then I asked him, “Have you considered the water and the moon? Water flows away but is never lost; the moon waxes and wanes, but neither increases nor diminished. If you look at its changing aspect, the universe passes in the twinkling of an eye; but if you look at its changeless aspect, all creatures including ourselves are imperishable. What reason have you to envy other things? Besides, everything in this universe has its owner; and if it does not belong to me not a tiny speck can I take. The sole exceptions are the cool breeze on the river, the bright moon over the hills. These serve as music to our ears, as colour to our eyes; these we can take freely and enjoy forever; these are inexhaustible treasures supplied by the Creator, and things in which we can delight together.
        \\
         \hline
        \textbf{Q} \begin{CJK}{UTF8}{gkai}
        下列对文段的理解和分析不正确的一项是
        \end{CJK}
         &\textbf{Q} Choose one from the following options which is incorrect understanding and analysis of the document.\\
        A. \begin{CJK}{UTF8}{gkai}
        主客对话的内容实际上是作者内心思想矛盾和感情复杂的反映，运用主客问答的方式，可以使行文波澜起伏，摇曳多姿，作者的思想感情也因此充分展现。
        \end{CJK} &
        A. The content of the dialogue between the author and his friend is actually a reflection of the author’s inner contradictions and complex feelings. The use of question-and-answer methods can make the writing ups and downs, swaying, and fully demonstrate the author’s thoughts and feelings.\\
        B. \begin{CJK}{UTF8}{gkai}
        作者通过多组对比，揭示出“悲”的原因。文中既写了曹孟德一世之雄的兴亡之悲，也写了由宇宙无穷与人生短暂的对比所生之悲，还写了现实与理想的矛盾所生之悲。
        \end{CJK} & 
        B. The author reveals the cause of "sorrow" through multiple groups of comparisons. The article not only writes the tragedy of the rise and fall of Cao Cao, but also the tragedy born from the contrast between the infinity of the universe and the short-lived life, and the tragedy born from the contradiction between reality and ideal.\\
        C. \begin{CJK}{UTF8}{gkai}
        作者以江水明月作比，说明世间万物和人生既有变的一面，也有不变的一面，阐述了自然万物变化与永恒的哲理，表现出作者无奈消极的人生态度。$\star$
        \end{CJK} & 
        \textbf{C. The author uses the river, water and the moon as a comparison to illustrate that everything in the world and life has both a changeable side and an unchanging side. He expounds the change and eternal philosophy of all things in nature, showing the author's helpless and passive attitude towards life.} $\star$\\
        D. \begin{CJK}{UTF8}{gkai}
        文段句式骈散结合，结构、句法、韵律都相对自由。大量对偶句的使用使文章参差错落，整齐简约，极富声韵之美。
        \end{CJK} &
        D. The paragraphs and sentences are combined with parallel and prose, and the structure, syntax, and rhythm are relatively free. The use of a large number of antithetical sentences makes the article jumbled, neat and simple, and full of beauty of rhyme and rhyme.\\
    \hline
    \end{tabular}
    \caption{An example of document in Classical Chinese with questions (left) and English translation (right), $\star$ is the correct option. And this question is an example of sentiment}
    \label{tab:classical_chinese}
\end{table*}

\begin{table*}[ht]
    \centering
    \begin{tabular}{m{13cm}}
    \toprule
    \begin{CJK}{UTF8}{gkai}
    故乡（节选）我这时很兴奋，但不知道怎么说才好，只是说：“阿！闰土哥，——你来了？……” 我接着便有许多话，想要连珠一般涌出：角鸡，跳鱼儿，贝壳，猹……但又总觉得被什么挡着似的，单在脑里面回旋，吐不出口外去。 他站住了，脸上现出欢喜和凄凉的神情；动着嘴唇，却没有作声。他的态度终于恭敬起来了，分明的叫道：“老爷！……” 我似乎打了一个寒噤；我就知道，我们之间已经隔了一层可悲的厚障壁了。我也说不出话。 他回过头去说，“水生，给老爷磕头。”便拖出躲在背后的孩子来，这正是一个廿年前的闰土，只是黄瘦些，颈子上没有银圈罢了。“这是第五个孩子，没有见过世面，躲躲闪闪……” 母亲和宏儿下楼来了，他们大约也听到了声音。 “老太太。信是早收到了。我实在喜欢的了不得，知道老爷回来……”闰土说。 “阿，你怎的这样客气起来。你们先前不是哥弟称呼么？还是照旧：迅哥儿。”母亲高兴的说。 “阿呀，老太太真是……这成什么规矩。那时是孩子，不懂事……”闰土说着，又叫水生上来打拱，那孩子却害羞，紧紧的只贴在他背后。 “他就是水生？第五个？都是生人，怕生也难怪的；还是宏儿和他去走走。”母亲说。 宏儿听得这话，便来招水生，水生却松松爽爽同他一路出去了。母亲叫闰土坐，他迟疑了一回，终于就了坐，将长烟管靠在桌旁。
    \end{CJK} \\
    \midrule
    \textbf{Translation:} \\
    \emph{My old home} (excerpt) \textbf{Lu xun} Delighted as I was, I did not know how to express myself, and could only say: "Oh! Jun-tu—so it's you? . . ."
    After this there were so many things I wanted to talk about, they should have poured out like a string of beads: woodcocks, jumping fish, shells, zha. . . . But I was tongue-tied, unable to put all I was thinking into words.
    He stood there, mixed joy and sadness showing on his face. His lips moved, but not a sound did he utter. Finally, assuming a respectful attitude, he said clearly:
    "Master! . . ."
    I felt a shiver run through me; for I knew then what a lamentably thick wall had grown up between us. Yet I could not say anything.
    He turned his head to call:
    "Shui-sheng, bow to the master." Then he pulled forward a boy who had been hiding behind his back, and this was just the Jun-tu of twenty years before, only a little paler and thinner, and he had no silver necklet.
    "This is my fifth," he said. "He's not used to company, so he's shy and awkward."
    Mother came downstairs with Hung-erh, probably after hearing our voices.
    "I got your letter some time ago, madam," said Jun-tu. "I was really so pleased to know the master was coming back. . . ."
    "Now, why are you so polite? Weren't you playmates together in the past?" said mother gaily. "You had better still call him Brother Hsun as before."
    "Oh, you are really too. . . . What bad manners that would be. I was a child then and didn't understand." As he was speaking Jun-tu motioned Shui-sheng to come and bow, but the child was shy, and stood stock-still behind his father.
    "So he is Shui-sheng? Your fifth?" asked mother. "We are all strangers, you can't blame him for feeling shy. Hung-erh had better take him Out to play."
    
    When Hung-eth heard this he went over to Shui-sheng, and Shui-sheng went out with him, entirely at his ease. Mother asked Jun-tu to sir down, and after a little hesitation he did so; leaning his long pipe against the table.\\
    \bottomrule

    \end{tabular}

    \caption{The document (top) of example in Table~6 and its English translation. (bottom)}
    \label{tab:hometown}
\end{table*}

\bibliography{anthology,custom}
\bibliographystyle{plain}

\clearpage
\section{Datasheet}
\subsection{Motivation}
\subsubsection*{For what purpose was the dataset created? Was there a specific task in mind? Was there a specific gap that needed to be filled? Please provide a description. } we seek to build a native-level Chinese comprehension system, we collect documents with questions from the exam questions for the Chinese course in China’s high schools, which are designed to evaluate the language proficiency of native Chinese youth. These questions are also not easy for native Chinese speakers and aim to push the frontier of building native-level Chinese MRC models.

\subsubsection*{Who created this dataset (e.g., which team, research group) and on behalf of which entity (e.g., company, institution, organization)? }
It was created by Haihua Institute for Frontier Information Technology.

\subsubsection*{Who funded the creation of the dataset? If there is an associated grant, please provide the name of the grantor and the grant name and number. }
Haihua Institute for Frontier Information Technology.

\subsection{Composition}

\subsubsection*{What do the instances that comprise the dataset represent (e.g., documents, photos, people, countries)? Are there multiple types of instances (e.g., movies, users, and ratings; people and interactions between them; nodes and edges)? Please provide a description. }

Each instance contains a document and a list of multiple-choice questions related to this document. And each question is comprised of a question text and 2~4 options, of which exactly one is correct, and the correct answer is also included.

\begin{table}[h!]
    \centering
    \caption{Fields and description in \name}
    \begin{tabular}{m{4cm} | m{9cm}}
    \hline
        Field & Description \\
    \hline
        ID & Document ID \\
        Content & Document text \\
        Questions & A list of quetions \\
        Questions:Question & Question text \\
        Qusetions:Answer & The ground truth answer \\\
        Questions:Q\_id	& Question id\\
    \hline
        Type (Annotated in validation/test set) & Writing style of document, 00 for moder-style (without poetry), 11 for classical-style (without poetry), 22 for classical poetry, 33 for modern poetry \\
    \hline
    \end{tabular}
    \label{tab:my_label}
\end{table}

\subsubsection*{How many instances are there in total (of each type, if appropriate)?}
NCR consists of 6315  documents with 15419 questions for training, 1000 documents with 2443 questions for validation and 1073 documents with 2615 questions for testing.

\subsubsection*{Does the dataset contain all possible instances or is it a sample (not necessarily random) of instances from a larger set? If the dataset is a sample, then what is the larger set? Is the sample representative of the larger set (e.g., geographic coverage)? If so, please describe how this representativeness was validated/verified. If it is not representative of the larger set, please describe why not (e.g., to cover a more diverse range of instances, because instances were withheld or unavailable). }
The dataset is a sample of instances from the larger set containing all exam questions for the Chinese course in China’s high schools. It is impossible to collect all the questions since there are too many questions online and many new questions will be generated every day.

\subsubsection*{What data does each instance consist of? “Raw” data (e.g., unprocessed text or images)or features? In either case, please provide a description. }

Data is all in the form of text.

\subsubsection*{Is there a label or target associated with each instance? If so, please provide a description. }
We provide each question with the correct answer. For data in the validation and test set, we also provide the writing style of the document.

\subsubsection*{Is any information missing from individual instances? If so, please provide a description, explaining why this information is missing (e.g., because it was unavailable). This does not include intentionally removed information, but might include, e.g., redacted text. }
There is no information missing.

\subsubsection*{Are relationships between individual instances made explicit (e.g., users’ movie ratings, social network links)? If so, please describe how these relationships are made explicit.}
A document is associated with several questions, these questions share the same document.

\subsubsection*{Are there recommended data splits (e.g., training, development/validation, testing)? If so, please provide a description of these splits, explaining the rationale behind them.}
We randomly split the dataset collected online at the document level, with 6315 for training, 1000 for validation and 1000 for testing. To make sure our test set has sufficient novel questions that never appear online, we also invited a few high-school Chinese teachers to manually generate 193 questions for a total of 73 additional documents to augment the test set. Finally NCR consists of 6315  documents with 15419 questions for training, 1000 documents with 2443 questions for validation and 1073 documents with 2615 questions for testing.  

\subsubsection*{Are there any errors, sources of noise, or redundancies in the dataset? If so, please provide a description.}
There are no errors, sources of noise, or redundancies in the dataset because we filter out these data.

\subsubsection*{Is the dataset self-contained, or does it link to or otherwise rely on external resources (e.g., websites, tweets, other datasets)? If it links to or relies on external resources, a) are there guarantees that they will exist, and remain constant, over time; b) are there official archival versions of the complete dataset (i.e., including the external resources as they existed at the time the dataset was created); c) are there any restrictions (e.g., licenses, fees) associated with any of the external resources that might apply to a future user? Please provide descriptions of all external resources and any restrictions associated with them, as well as links or other access points, as appropriate.}
The dataset is self-contained.

\subsubsection*{Does the dataset contain data that might be considered confidentiality, data that includes the content of individuals' non-public communications)? If so, please provide a description.}
No, all data is public.

\subsubsection*{Does the dataset contain data that, if viewed directly, might be offensive, insulting, threatening, or might otherwise cause anxiety? If so, please describe why.}
The dataset does not contain any data that may be offensive, insulting, threatening, or might otherwise cause anxiety.

\subsubsection*{Does the dataset relate to people? If not, you may skip the remaining questions in this subsection.}
This dataset does not relate to people.

\subsection{Collection}
\subsubsection*{How was the data associated with each instance acquired? Was the data directly observable (e.g., raw text, movie ratings), reported by subjects (e.g., survey responses), or indirectly inferred/derived from other data (e.g., part-of-speech tags, model-based guesses for age or language)? If data was reported by subjects or indirectly inferred/derived from other data, was the data validated/verified? If so, please describe how. }
A document is associated with several questions, these questions share the same document, which is directly observable.

\subsubsection*{What mechanisms or procedures were used to collect the data (e.g., hardware apparatus or sensor, manual human curation, software program, software API)? How were these mechanisms or procedures validated?}
The data was collected by manual human curation. And the collected data was validated by another person to ensure the quality.

\subsubsection*{Who was involved in the data collection process (e.g., students, crowdworkers, contractors) and how were they compensated (e.g., how much were crowdworkers paid)?}
\begin{CJK}{UTF8}{gkai}
We contracted out the data collection process to 海天瑞声 ( SpeechOcean, http://en.speechocean.com/)
\end{CJK}

\subsubsection*{Over what timeframe was the data collected? Does this timeframe match the creation time frame of the data associated with the instances (e.g., recent crawl of old news articles)? If not, please describe the timeframe in which the data associated with the instances was created.}
The data collection process lasted for around 40 days. Since the dataset contains many articles in classical Chinese, which can be date back to thousands of years ago.

\subsubsection*{Were any ethical review processes conducted (e.g., by an institutional review board)? If so, please provide a description of these review processes, including the outcomes, as well as a link or other access point to any supporting documentation. }
No ethical review processes were conducted.

\subsection{Processing}
\subsubsection*{Was any preprocessing/cleaning/labeling of the data done (e.g., discretization or bucketing, tokenization, part-of-speech tagging, SIFT feature extraction, removal of instances, processing of missing values)? If so, please provide a description. If not, you may skip the remainder of the questions in this subsection.}
We clean the data by filter out some questions which are base on the format. For example, some questions are about the marked or boldface words.

\subsubsection*{Was the “raw” data saved in addition to the preprocessed/cleaned/labeled data (e.g., to support unanticipated future uses)? If so, please provide a link or other access point to the “raw” data. }
No.

\subsubsection*{Is the software used to preprocess/clean/label the instances available? If so, please provide a link or other access point. }
No.

\subsection{Uses}
\subsubsection*{Has the dataset been used for any tasks already? If so, please provide a description.}
To examine the limit of current MRC methods, we organized a 3-month-long online competition using NCR with a training and validation set released.  Participants are allowed to use any open-access pre-trained model or any open-access unlabeled data.  Use of any external MRC supervision is forbidden, since a portion of the test questions are possibly accessible online. This aims to prevent human annotations overlapping with our held-out data for a fair competition. There are a total of 141 participating teams and the best submission model with the highest test accuracy is taken as the competition model. The team is from an industry lab. They first pre-trained an XLNet-based model on a company-collected large corpus. For each question, they use an information retrieval tool Okapi BM25 to extract the most relevant parts from the document and then run this pre-trained model for answer selection based on the extracted texts. The final model with the highest accuracy is released: \url{https://github.com/xssstory/NCR_competition_model}

\subsubsection*{Is there a repository that links to any or all papers or systems that use the dataset?}
The competition website: \url{https://www.biendata.xyz/competition/haihua_2021/}

Github for the competition model with the highest accuracy: \url{https://github.com/xssstory/NCR_competition_model}

Github for baselines:
\url{https://github.com/xssstory/NCR_baseline}

\subsubsection*{What (other) tasks could the dataset be used for?}
This dataset may also be used for a question-answering generation task. 
\subsubsection*{Is there anything about the composition of the dataset or the way it was collected and preprocessed/cleaned/labeled that might impact future uses? For example, is there anything that a future user might need to know to avoid uses that could result in unfair treatment of individuals or groups (e.g., stereotyping, quality of service issues) or other undesirable harms (e.g., financial harms, legal risks) If so, please provide a description. Is there anything a future user could do to mitigate these undesirable harms?}
Users should keep in mind that the questions and their answer comes from different teachers. 

\subsubsection*{Are there tasks for which the dataset should not be used? If so, please provide a description.}
Unknown.

\subsection{Distribution}
\subsubsection*{Will the dataset be distributed to third parties outside of the entity (e.g., company, institution, organization) on behalf of which the dataset was created? }
Yes, the dataset is public now.

\subsubsection*{How will the dataset will be distributed (e.g., tarball on website, API, GitHub)? Does the dataset have a digital object identififier (DOI)? }
The dataset is available at \url{https://github.com/xssstory/NCR_competition_model} and \url{https://drive.google.com/drive/folders/1Ci-KLHKk-yP-y5fWX4_cU8bA2fL_q76e?usp=sharing}

\subsubsection*{When will the dataset be distributed?}
It is available now.

\subsubsection*{Will the dataset be distributed under a copyright or other intellectual property (IP) license, and/or under applicable terms of use (ToU)? If so, please describe this license and/or ToU, and provide a link or other access point to, or otherwise reproduce, any relevant licensing terms or ToU, as well as any fees associated with these restrictions.}
This dataset is released under the CC BY-SA 4.0 license for general research purposes.

\subsubsection*{Have any third parties imposed IP-based or other restrictions on the data associated with the instances? If so, please describe these restrictions, and provide a link or other access point to, or otherwise reproduce, any relevant licensing terms, as well as any fees associated with these restrictions.}
No.

\subsubsection*{Do any export controls or other regulatory restrictions apply to the dataset or to individual instances? If so, please describe these restrictions, and provide a link or other access point to, or otherwise reproduce, any supporting documentation.}
No

\subsection{Maintance}
\subsubsection*{Who is supporting/hosting/maintaining the dataset?}
The dataset will be hosted on GitHub and google drive and will be maintained by Shusheng Xu and Yichen Liu.

\subsubsection*{How can the owner/curator/manager of the dataset be contacted (e.g., email address)?}
The maintainers can be contacted at xuss20@mails.tsinghua.edu.cn and y17043@nyu.edu

\subsubsection*{Is there an erratum? If so, please provide a link or other access point.}
No.

\subsubsection*{Will the dataset be updated (e.g., to correct labeling errors, add new instances, delete instances)? If so, please describe how often, by whom, and how updates will be communicated to users (e.g., mailing list, GitHub)?}
The dataset will be updated if necessary, updates will be communicated via the project website \url{https://sites.google.com/view/native-chinese-reader/}  Github at \url{https://github.com/xssstory/NCR_competition_model} and \url{https://github.com/xssstory/NCR_baseline}

\subsubsection*{If the dataset relates to people, are there applicable limits on the retention of the data associated with the instances (e.g., were individuals in question told that their data would be retained for a fixed period of time and then deleted)? If so, please describe these limits and explain how they will be enforced.}
N/A

\subsubsection*{Will older versions of the dataset continue to be supported/hosted/maintained? If so, please describe how. If not, please describe how its will be communicated to users.}
Yes, it will be communicated to users via the project website \url{https://sites.google.com/view/native-chinese-reader/}  Github at \url{https://github.com/xssstory/NCR_competition_model} and \url{https://github.com/xssstory/NCR_baseline}

\subsubsection*{If others want to extend/augment/build on/contribute to the dataset, is there a mechanism for them to do so? If so, please provide a description. Will these contributions be validated/verified? If so, please describe how. If not, why not? Is there a process for communicating/distributing these contributions to other users? If so, please provide a description. }
This is not supported now.